\documentclass[sigconf]{acmart}     % 

\usepackage{tabularx}
\usepackage{booktabs}
\usepackage{multirow}
\usepackage{graphicx}
\usepackage{amsmath}
\usepackage{listings}
\usepackage{xspace}
\usepackage{makecell}

\setcopyright{acmlicensed}
\copyrightyear{2025}
\acmYear{2025}
\acmDOI{XXXXXXX.XXXXXXX}
\acmISBN{978-1-4503-XXXX-X/25/07}
\acmConference[SIGIR FinIR'25]{The 2nd Workshop on Financial Information Retrieval in the Era of Generative AI}{July 17, 2025}{Padua, Italy}

\title{Structuring the Unstructured: A Multi-Agent System for Extracting and Querying Financial KPIs and Guidance}

\author{Chanyeol Choi}
\affiliation{
  \institution{LinqAlpha}
  \city{New York}
  \state{NY}
  \country{USA}
}

\author{Alejandro Lopez-Lira}
\affiliation{
  \institution{University of Florida}
  \city{Gainesville}
  \state{FL}
  \country{USA}
}

\author{Yongjae Lee}
\affiliation{
  \institution{UNIST}
  \city{Ulsan}
  \country{Republic of Korea}
}

\author{Jihoon Kwon}
\affiliation{
  \institution{LinqAlpha}
  \city{New York}
  \state{NY}
  \country{USA}
}

\author{Minjae Kim}
\affiliation{
  \institution{LinqAlpha}
  \city{New York}
  \state{NY}
  \country{USA}
}

\author{Juneha Hwang}
\affiliation{
  \institution{LinqAlpha}
  \city{New York}
  \state{NY}
  \country{USA}
}

\author{Minsoo Ha}
\affiliation{
  \institution{LinqAlpha}
  \city{New York}
  \state{NY}
  \country{USA}
}

\author{Chaewoon Kim}
\affiliation{
  \institution{LinqAlpha}
  \city{New York}
  \state{NY}
  \country{USA}
}

\author{Jaeseon Ha}
\affiliation{
  \institution{LinqAlpha}
  \city{New York}
  \state{NY}
  \country{USA}
}

\author{Suyeol Yun}
\affiliation{
  \institution{LinqAlpha}
  \city{New York}
  \state{NY}
  \country{USA}
}

\author{Jin Kim}
\affiliation{
  \institution{LinqAlpha}
  \city{New York}
  \state{NY}
  \country{USA}
}

\renewcommand{\shortauthors}{Choi et al.}

\begin{document}

\begin{abstract}
Extracting and retrieving structured and quantitative insights from unstructured financial filings is essential in investment research, yet remains time-consuming and resource-intensive.
Conventional approaches in practice rely heavily on labor-intensive manual processes, limiting scalability and delaying the research workflow.
In this paper, we propose an efficient and scalable method for both \emph{extracting} and \emph{retrieving} financial insights accurately, leveraging a multi-agent system composed of large language models.
Our proposed multi-agent system consists of two specialized agents: the \emph{Extraction Agent} and the \emph{Text-to-SQL Agent}.
The \textit{Extraction Agent} automatically identifies key performance indicators from unstructured financial text, standardizes their formats, and verifies their accuracy.
On the other hand, the \textit{Text-to-SQL Agent} generates executable SQL statements from natural language queries, allowing users to access structured data accurately without requiring familiarity with the database schema.
Through experiments, we demonstrate that our proposed system effectively transforms unstructured text into structured data accurately and enables precise retrieval of key information.
First, we demonstrate that our system achieves approximately 95\% accuracy in transforming financial filings into structured data, matching the performance level typically attained by human annotators.
Second, in a human evaluation of the retrieval task -- where natural language queries are used to search information from structured data -- 91\% of the responses were rated as correct by human evaluators.
In both evaluations, our system generalizes well across financial document types, consistently delivering reliable performance.
\end{abstract}


\maketitle

\section{Introduction}

Investment research depends critically on the analysis of unstructured corporate disclosures.
In practice, analysts often extract key performance indicators (KPIs) either by manually scanning lengthy documents or by locating figures scattered across external sources and internal summaries.
As firms continue to release a growing volume of earnings transcripts, financial reports, and regulatory filings, such manual processes have become increasingly time-consuming and impractical.
To ensure both timeliness and breadth of coverage, scalable methods for extracting and retrieving insights from these texts are now essential to informed decision-making.
Recent advances in large language models (LLMs) have opened new possibilities for processing unstructured documents at scale, offering a promising foundation for automating financial information \emph{extraction} and \emph{retrieval}.
These models can, in principle, transform complex textual data into structured insights and support natural language interactions with document content.

However, despite their potential, LLMs yet remain unreliable in real-world financial workflows.
When used in isolation, they often misinterpret numeric guidance, extract semantically incorrect values, or omit critical metrics altogether.
Such failures are largely due to limitations in layout sensitivity and the difficulty of distinguishing factual signals from surrounding narrative.
To address these shortcomings, recent work has explored multi-agent frameworks, in which multiple LLMs collaborate to validate, cross-check, and refine each other’s outputs—improving accuracy, reducing hallucinations, and enhancing reliability in financial document analysis.

In this context, we propose an efficient and scalable method for both \emph{extracting} and \emph{retrieving} financial insights accurately, leveraging a multi-agent system composed of LLMs.
Our system consists of two specialized agents: an \emph{Extraction Agent}, which parses and validates KPIs using domain-tuned prompts and logic, and a \emph{Text-to-SQL Agent}, which translates analyst queries into executable SQL over the structured output.
By explicitly decomposing the workflow, the system allows targeted evaluation, modular tuning, and the integration of feedback loops at each stage-enabling iterative refinement and robust error handling.
This modular design not only automates manual workflows but also ensures both high-fidelity structuring and precise, trustworthy retrieval. Evaluated on a diverse set of SEC filings and real analyst tasks, our system demonstrates significant improvements in accuracy, controllability, and scalability over traditional and monolithic LLM baselines.

\section{Related Work}

\subsection{Financial Information Extraction}

Extracting structured data from financial documents such as 10-K, 10-Q, and earnings call transcripts has been a long-standing challenge. Early approaches employed rule-based parsing and template-matching techniques to identify financial entities and events, but these methods tend to be brittle under formatting variation or narrative complexity~\cite{sheikh2012rule,costantino2008information}. Supervised event-based extraction systems have also been explored, such as Malik et al.'s classifier-based framework for quantitative financial events~\cite{malik2011accurate} and SENTiVENT, which provides fine-grained labeled corpora for economic news event extraction~\cite{jacobs2022sentivent}. While these systems demonstrated strong performance within narrow domains, they require substantial feature engineering and lack generalizability across document types.

More recently, LLM-based extractors have been introduced to capture KPIs in a zero- or few-shot manner~\cite{costantino1997nlp}. While promising, these models often lack domain grounding and tend to hallucinate values or overlook subtle context such as fiscal guidance, non-GAAP metrics, or period ranges. In contrast, our system explicitly integrates domain-specific logic and validation steps into the extraction pipeline to ensure reliability and robustness across diverse financial disclosures.

\subsection{LLMs and Financial Document Understanding}

Large language models (LLMs) have been increasingly applied to financial document tasks such as summarization, question answering, and data extraction~\cite{yang2024evaluating, li2024extracting, shah2024multi, ziegler2024automating, srivastava2024evaluating}. While they demonstrate promising capabilities in processing general text, recent studies highlight fundamental limitations when applied to semi-structured or domain-specific financial content.

For example, \citet{srivastava2024evaluating} find that LLMs struggle with arithmetic reasoning in financial QA, particularly when aggregating values across tables or interpreting fiscal context. Similarly, Ziegler~\cite{ziegler2024automating} observes that LLMs lack consistent understanding of financial and environmental terminology, often requiring external schema knowledge to extract accurate insights. Shah et al.~\cite{shah2024multi} further note that despite their potential, current models fail to maintain consistency across multi-document financial QA, particularly when referencing nuanced company disclosures.

Moreover, hallucination, factual drift, and inability to trace predictions to source spans remain persistent issues~\cite{li2024extracting}. These challenges are compounded when LLMs are applied in a single-pass, end-to-end setting, where intermediate errors cannot be audited or corrected.

Our system addresses these limitations by separating responsibilities across two dedicated agents. The Extraction Agent ensures grounded, validated KPI outputs using financial logic, while the Text-to-SQL Agent enables structured querying over these results. This modularity allows iterative refinement and greater reliability, addressing key shortcomings identified in recent LLM-based financial IR studies.

\subsection{Text-to-SQL and Financial Querying}

Text-to-SQL systems aim to translate natural language questions into executable SQL over structured data. While general-purpose models have advanced considerably with benchmarks like Spider~\cite{singh2025survey}, their effectiveness in financial domains remains limited. Challenges arise from ambiguous query phrasing, implicit units or timeframes, and domain-specific schema structures~\cite{wretblad2024bridging, zhang2024finsql}.

To address this, recent studies have introduced financial-specific Text-to-SQL datasets such as FinSQL~\cite{zhang2024finsql} and BookSQL~\cite{kumar2024booksql}, which provide training corpora grounded in accounting or financial reporting. However, even with domain adaptation, models often produce syntactically valid but semantically incorrect queries-such as mixing YoY and QoQ values, misidentifying column types, or violating fiscal constraints~\cite{song2024enhancing, wretblad2024bridging}.

Our framework mitigates these issues by embedding schema-aware prompting, constraint validation, and semantic consistency checks into a dedicated Text-to-SQL Agent. This modular design allows fine-grained control over unit interpretation, time alignment, and execution correctness-enabling robust, analyst-friendly query interfaces for dynamic financial datasets.

\subsection{Multi-Agent Architectures for Modular Reasoning}

Recent advances in LLM-based multi-agent systems have shown that task decomposition and agent coordination can significantly improve reasoning robustness, modularity, and scalability~\cite{tran2025multiagent, li2024survey, agashe2023coordination}. Rather than relying on a single-shot monolithic LLM pipeline, multi-agent frameworks assign subtasks to specialized agents with explicit responsibilities and interaction protocols. This structure allows intermediate outputs to be validated, debugged, and iteratively improved.

For example, AgentNet~\cite{yang2025agentnet} explores decentralized architectures for coordinating LLM agents in complex planning settings, while LLM-Coordination~\cite{agashe2023coordination} analyzes how cognitive structures such as Theory of Mind can enhance collaborative task completion. Surveys by Li et al.~\cite{li2024survey} and Tran et al.~\cite{tran2025multiagent} emphasize that multi-agent designs offer benefits such as reduced hallucination, explicit memory sharing, and better long-range task alignment.

\begin{figure*}[t]
    \centering
    \includegraphics[width=0.8\textwidth]{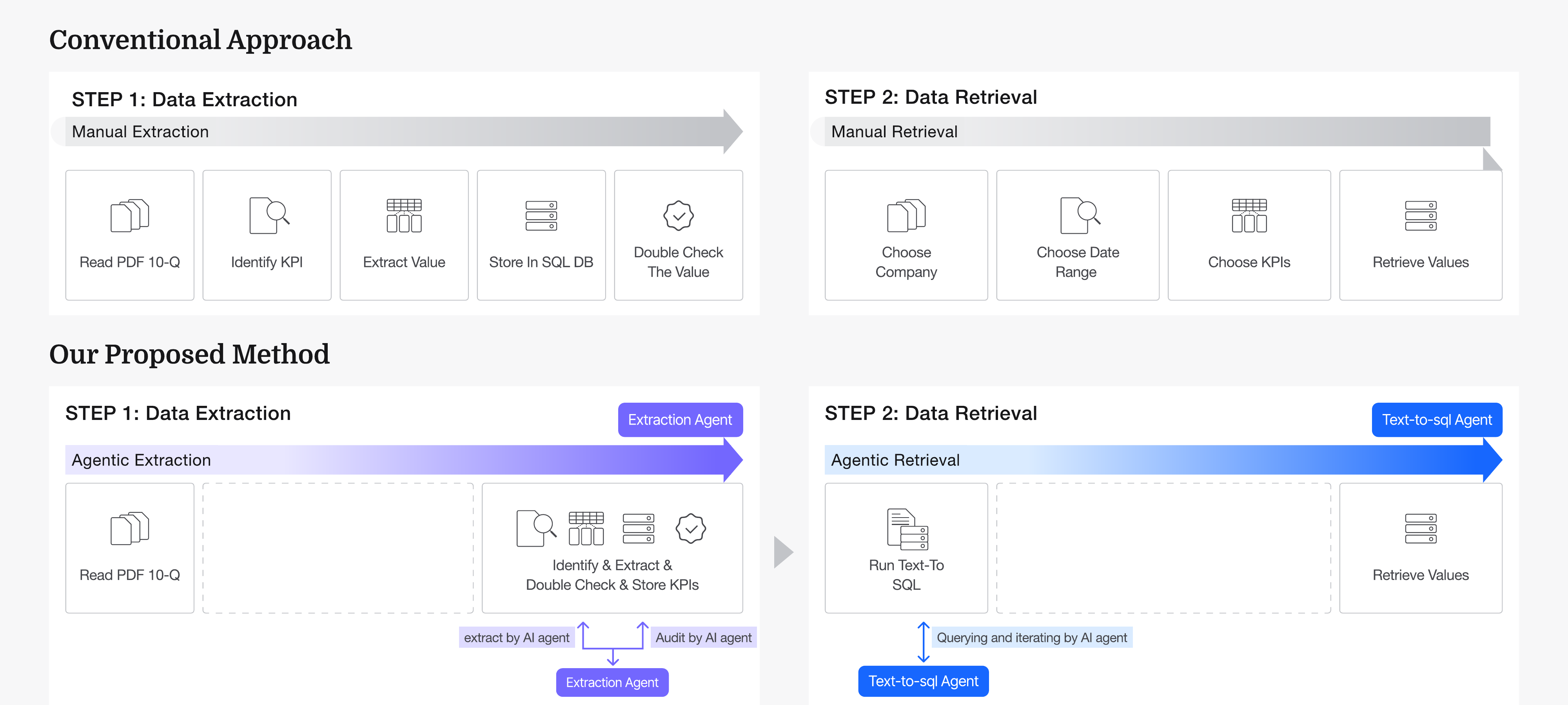}
    \caption{Comparison between the conventional manual pipeline and our proposed agentic system for financial KPI extraction and retrieval. While the conventional approach (Top) relies on labor-intensive human processes across both extraction and retrieval steps, our method (Bottom) automates the workflow using two agents—\emph{Extraction Agent} and \emph{Text-to-SQL Agent}—which collaborate to deliver scalable, accurate, and end-to-end structuring and querying of financial data.}
    \Description{Diagram comparing manual and agentic systems for financial KPI extraction and retrieval. The manual system involves human steps, while the agentic system uses two collaborating agents for automation.}
    \label{fig:method_comparison}
\end{figure*}

\section{Method}

This section introduces our agentic system for automated financial KPI extraction and querying. We begin by outlining the challenges of manual workflows that analysts face when working with financial documents. We then present a modular two-agent architecture consisting of (1) an \textit{Extraction Agent}, which transforms raw filings into structured KPI records, and (2) a \textit{Text-to-SQL Agent}, which enables natural language querying over these records. Finally, we describe the technical architecture that supports scalable deployment and integration.

\subsection{Motivation: Manual KPI Extraction is Inefficient}

\begin{figure}[h]
    \centering
    \includegraphics[width=0.45\textwidth]{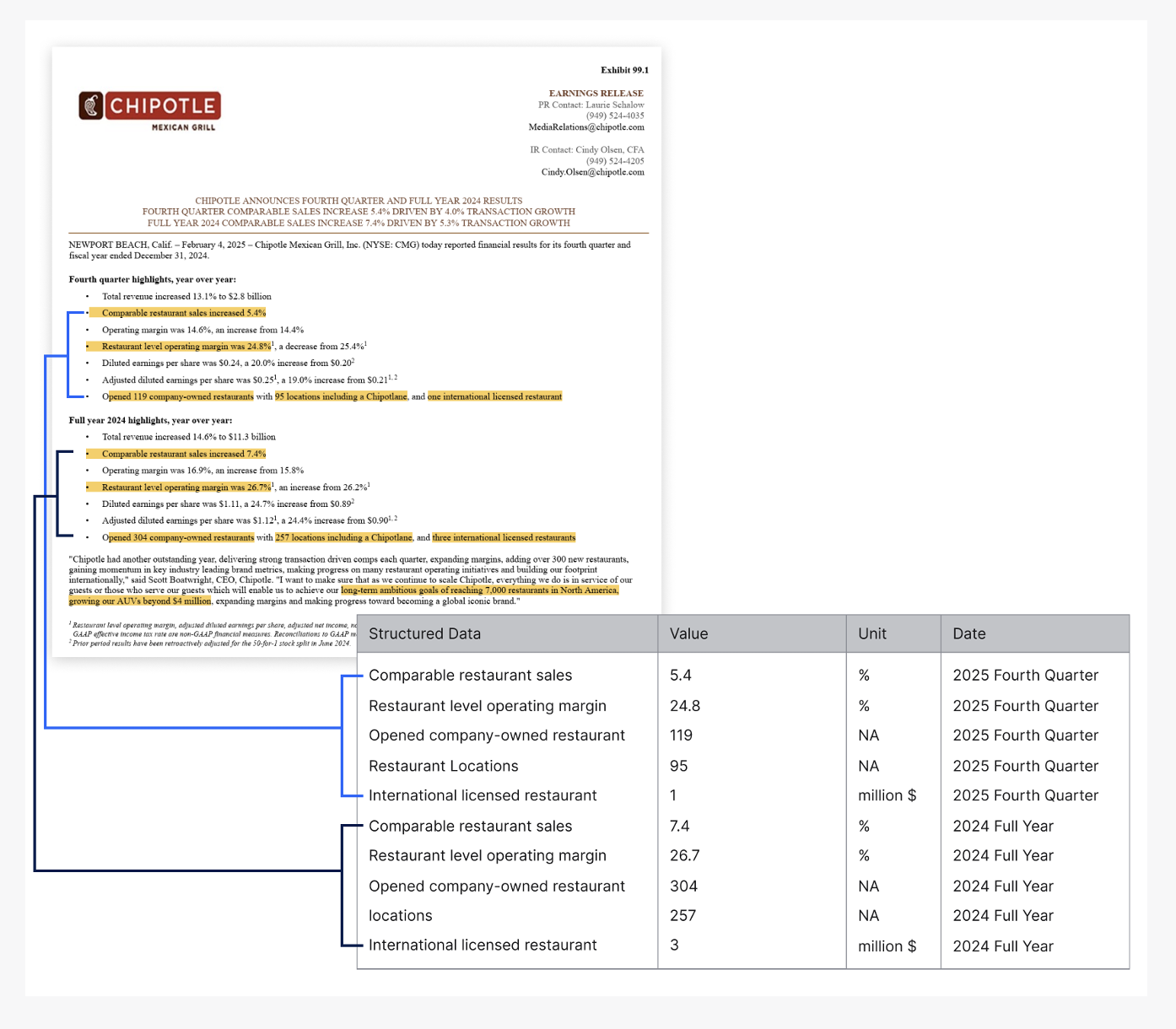}
    \caption{Manual KPI extraction process from earnings releases (e.g., Chipotle). Analysts must parse text, resolve numeric ranges, and enter structured data manually.}
    \label{fig:manual_example}
\end{figure}

Financial analysts frequently extract key performance indicators (KPIs) from documents like 10-Ks, 10-Qs, earnings releases, and investor transcripts. As illustrated in Fig.~\ref{fig:manual_example}, this process involves manually identifying relevant statements, interpreting financial terms, normalizing figures (e.g., converting "22–24\%" to midpoint 23\%), and transferring them to downstream tools like spreadsheets or databases. This pipeline is slow, error-prone, and often lacks standardization—especially when dealing with ranges, units, and ambiguous time periods.

\subsection{System Overview: A Two-Agent Architecture}

To address these inefficiencies, we propose a modular system based on a two-step multi-agent architecture, shown in Fig.~\ref{fig:system_architecture}. The system separates the extraction and querying components into specialized agents, each designed to optimize for financial accuracy, controllability, and interpretability.

The pipeline proceeds as follows:
\begin{enumerate}
    \item \textbf{Document Saving}: Raw inputs including PDFs, HTML earnings pages, and transcripts are stored.
    \item \textbf{Layout \& Parsing}: Documents are OCR-processed (if necessary) and segmented by logical sections.
    \item \textbf{Preprocessing}: Text is normalized; tables and numeric spans are detected.
    \item \textbf{Embedding \& Structuring}: Domain-specific embedding models identify KPI-relevant segments.
    \item \textbf{Step 1 – Extraction Agent}: KPIs, fiscal periods, and metadata are extracted and validated.
    \item \textbf{Step 2 – Text-to-SQL Agent}: Analyst queries are translated into executable SQL via a schema-aware model.
\end{enumerate}

\begin{figure*}[t]
    \centering
    \includegraphics[width=0.9\linewidth]{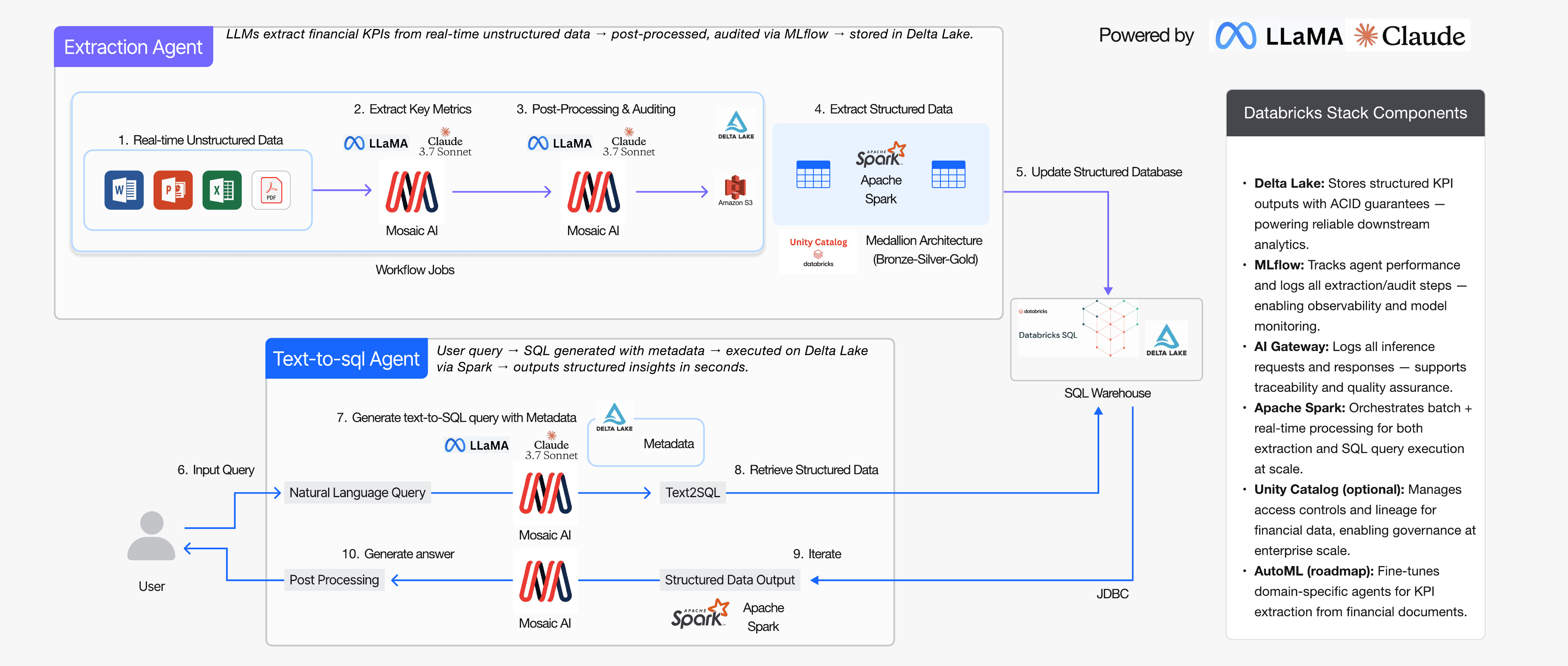}
    \caption{Overview of our multi-agent system for financial document understanding. The \emph{Extraction Agent} processes unstructured documents to extract structured KPIs using LLMs (LLaMA, Claude) and stores them in Delta Lake via Apache Spark. The \emph{Text-to-SQL Agent} enables natural language querying by generating metadata-aware SQL queries, retrieving results through Databricks SQL. The system is built on the Databricks stack, including MLflow for auditing and AI Gateway for traceability.}
    \label{fig:extraction_evaluation}
\end{figure*}

\subsection{Extraction Agent}

The Extraction Agent transforms raw financial documents into structured, machine-readable KPI records through a seven-stage process combining neural reasoning and rule-based logic. The top side of Fig.~\ref{fig:extraction_evaluation} shows its detailed steps.

\begin{itemize}
    \item \textbf{Raw Data Parsing:} Layout-aware parsers segment filings into logical sections. OCR is applied where necessary.
    \item \textbf{Entity and Metric Detection:} Domain-tuned prompting identifies candidate KPI spans using span-level and taxonomy-aware templates.
    \item \textbf{Contextual Span Extraction:} Surrounding context like time periods and qualifiers (e.g., “guidance”, “non-GAAP”) is extracted alongside numeric values.
    \item \textbf{Preliminary Structuring:} KPI spans are mapped into \texttt{\{metric, value, unit, period, qualifier\}} schema.
    \item \textbf{Domain Rule Injection:} Heuristics normalize ranges, resolve units, disambiguate fiscal periods, and handle adjusted metrics.
    \item \textbf{QA-Driven Validation:} Consistency is verified via targeted questions; inconsistent outputs are flagged or corrected.
    \item \textbf{Final Structuring and Output:} Records are emitted as JSON with document metadata and confidence scores.
\end{itemize}

\subsection{Text-to-SQL Agent}

The Text-to-SQL Agent enables analysts to pose natural language questions over the structured KPI database. It translates queries like “What was the Q4 operating margin?” into executable SQL, handling financial semantics and schema variation. The bottom side of Fig.~\ref{fig:extraction_evaluation} shows its detailed steps.

\begin{itemize}
    \item \textbf{Query Understanding:} Parse target metrics, time constraints, aggregation logic, and qualifiers.
    \item \textbf{Schema Augmentation:} Inject prompt context with schema metadata, column types, aliases, and sample rows.
    \item \textbf{SQL Generation:} Use LLMs (e.g., Claude 3.7, GPT-4) with financial prompting and few-shot examples.
    \item \textbf{Constraint Validation:} Check syntax, unit consistency, temporal alignment, and qualifier accuracy.
    \item \textbf{Execution Feedback:} Test the query and rerank or regenerate if the output is implausible or empty.
    \item \textbf{Explanation Generation:} Translate SQL into natural language summaries to support interpretability.
\end{itemize}

\subsection{Overall Pipeline}

The overall pipeline is implemented on Databricks and supports scalable, auditable financial document processing.

\begin{itemize}
    \item \textbf{Document Ingestion:} Secure API/file-drop ingestion of filings into a unified data lake.
    \item \textbf{Parsing and Preprocessing:} Layout-aware parsing and normalization via PDF/HTML/OCR pipelines.
    \item \textbf{LLM Orchestration:} Claude 3.7 is used for structured extraction through prompt-based workflows.
    \item \textbf{Data Storage:} Validated KPI records are stored in Delta Lake tables with schema guarantees.
    \item \textbf{Text-to-SQL Interface:} Analysts query data using natural language interfaces mapped to the current schema.
    \item \textbf{Monitoring:} Rule-based and MLFlow-based tracking ensures reliability, version control, and auditability.
\end{itemize}

This architecture supports extensibility across new document types, robustness to schema drift, and integration with third-party systems or human-in-the-loop workflows.

\section{Experiments}

\renewcommand{\arraystretch}{1.3}
\begin{table*}[h]
\centering
\caption{Comparison of overall pipeline (Extraction Agent \& Text-to-SQL Agent) across configurations}
\label{tab:evaluation_results}
\begin{tabularx}{\linewidth}{
  >{\raggedright\arraybackslash}X
  >{\centering\arraybackslash}X
  >{\centering\arraybackslash}X
  >{\centering\arraybackslash}X
  >{\centering\arraybackslash}X
}
\toprule
\textbf{Config.} & \textbf{Extraction F1} & \textbf{Query Accuracy} & \textbf{Unit Error Rate} & \textbf{Throughput Gain} \\
\midrule
\texttt{GPT-4o} (naive zero-shot)        & 72.4\%         & 81.3\%         & 18.5\%        & 1.0× \\
\texttt{Claude-3.7-Sonnet} (domain prompt)     & 84.7\%         & 89.6\%         & 7.2\%         & 1.5× \\
\texttt{Claude-3.7-Sonnet} + Rule Injection & \textbf{91.2\%} & \textbf{94.8\%} & \textbf{2.3\%} & \textbf{2.4×} \\
\bottomrule
\end{tabularx}
\end{table*}

\subsection{Experimental Setup}

To evaluate the effectiveness of our multi-agent framework, we simulate real-world analyst workflows involving the extraction of KPIs from SEC filings (10-Ks, 8-Ks, earnings releases) and natural language querying over those KPIs. Our evaluation uses:
\begin{itemize}
    \item \textbf{Documents:} 1,200 SEC filings from publicly traded companies.
    \item \textbf{Queries:} 312 natural language questions derived from hedge fund analyst logs.
\end{itemize}

Each document was manually annotated by trained financial analysts to identify ground-truth KPIs, including revenue, operating income, free cash flow, and forward-looking guidance.
Annotations were cross-validated for consistency and aligned with GAAP definitions where applicable.
We compare three configurations:
\begin{itemize}
    \item \textbf{Baseline A (GPT-4 Zero-Shot):} Direct prompting without schema or constraints.
    \item \textbf{Baseline B (Claude 3.7 + Domain Prompt):} Structured prompting with schema knowledge.
    \item \textbf{Ours (Claude 3.7 + Rule Injection):} Full pipeline with rule-based validation and semantic constraints.
\end{itemize}

\begin{table}[t]
\centering
\caption{Performance on individual agent: \emph{Extraction Agent}}
\label{tab:individual_extraction}
\begin{tabular}{lc}
\toprule
\textbf{Metric} & \textbf{Value} \\
\midrule
Metric Precision         & 95.3\% \\
Structuring Accuracy     & 91.6\% \\
Unit Error Rate          & 2.3\% \\
QA Match Rate            & 92.3\% \\
Schema Compliance        & 100.0\% \\
\bottomrule
\end{tabular}
\end{table}

\subsection{Component-Level Evaluation and End-to-End Pipeline Comparison}

We first assess each agent in isolation to measure its contribution to overall accuracy in Table~\ref{tab:individual_extraction} and ~\ref{tab:individual_generation}.
Next, we evaluate the combined system across the three configurations in Table~\ref{tab:evaluation_results}. The results show that structured prompting and rule-based validation significantly improve both extraction and query accuracy while reducing error rates and increasing throughput.

\paragraph{Prompt Engineering Impact.}  
Domain-specific prompt tuning (Claude 3.7) improves extraction F1 by +12.3 points over GPT-4 and reduces unit error rate by 60\%, demonstrating the value of schema-aware prompting.

\paragraph{Rule Injection Effect.}  
Adding validation and heuristics significantly reduces semantic and numeric misalignment, especially for period mismatches and GAAP vs. non-GAAP misinterpretation.

\paragraph{Query Reliability.}  
Constraint-validated SQL generation yields a +13.5 point gain in query execution accuracy compared to zero-shot baselines, reflecting better alignment between user intent and schema-constrained SQL.

\paragraph{System Efficiency.}  
Our full framework achieves a 2.4× throughput gain by minimizing error propagation and reducing manual normalization overhead.

\begin{table}[t]
\centering
\caption{Performance on individual agent: \emph{Text-to-SQL Agent}}
\label{tab:individual_generation}
\begin{tabular}{lc}
\toprule
\textbf{Metric} & \textbf{Value} \\
\midrule
Intent Classification Accuracy   & 95.6\% \\
SQL Syntax Validity              & 98.2\% \\
Semantic Constraint Pass Rate    & 91.3\% \\
Execution Success Rate           & 96.4\% \\
Top-1 SQL Accuracy (Human Eval)  & 91.2\% \\
Explanation Relevance            & 90.5\% \\
\bottomrule
\end{tabular}
\end{table}

\subsection{Error Case Examples}

\begin{itemize}
    \item \textbf{GPT-4 Zero-Shot:} Interpreted “operating margin guidance of 15–17\%” as actual Q4 value (16.0\%).
    \item \textbf{Claude (no rule):} Extracted 16.0\% correctly but failed to flag it as guidance.
    \item \textbf{Claude + Rules:} Identified range as guidance and excluded it from GAAP-specific queries.
\end{itemize}

Our modular, rule-enhanced multi-agent architecture enables robust, interpretable financial QA. By decomposing the pipeline into extraction, validation, and SQL generation, we gain control over accuracy, explainability, and system-level efficiency. These improvements are particularly valuable for real-world analyst workflows, where numeric precision and auditability are paramount.

\section{Conclusion}

We presented a modular, multi-agent framework for extracting and querying structured information from unstructured financial documents such as 10-Ks, 8-Ks, and earnings transcripts. By decomposing the pipeline into a dedicated Extraction Agent and a Text-to-SQL Agent, our system addresses key challenges in financial information retrieval-including document layout variation, unit normalization, and semantic ambiguity in natural language queries.

Through prompt engineering and domain-specific validation, our architecture outperforms baseline LLM setups in both accuracy and robustness, enabling analysts to access high-fidelity KPI data with significantly reduced manual effort. The framework’s separation of concerns also allows for targeted tuning, iterative refinement, and easier auditability.

Future work includes expanding support to multilingual filings, integrating richer temporal logic, and enabling cross-document reasoning over longer time horizons. We believe this approach represents a scalable, trustworthy foundation for deploying generative AI in financial research workflows.

\appendix

\section{Prompt Engineering Examples}

This section compares two prompting strategies used in the Extraction Agent: a naive zero-shot prompt and a domain-aware prompt that incorporates schema, unit normalization, and period anchoring.

\begin{table}[h]
\centering
\caption{Prompt Engineering Comparison: Zero-Shot vs Domain-Aware}
\label{tab:prompt_comparison}
\renewcommand{\arraystretch}{1.3}
\begin{tabularx}{\linewidth}{>{\raggedright\arraybackslash}X >{\raggedright\arraybackslash}X}
\toprule
\textbf{Zero-Shot Prompt} & \textbf{Domain-Aware Prompt} \\
\midrule
\textbf{Prompt} \newline
\small \texttt{Extract key financial figures from the text below.} 
& 
\textbf{Prompt} \newline
\small \texttt{Extract financial KPIs from the earnings text using the schema below. Output in JSON. Normalize units and link metrics to fiscal periods.} \\

\textbf{Text} \newline
\small \texttt{In Q1 2024, revenue grew 12\% YoY to \$4.3 billion, beating consensus by \$150 million.} 
& 
\textbf{Text} \newline
\small \texttt{Same as left. Schema includes Period, Revenue (B), Revenue\_YoY\_Growth, and Consensus\_Delta (M).} \\

\textbf{Output (Expected)} \newline
\small \texttt{\{ Revenue: "4.3B", Growth: "12\%", Period: "Q1 2024" \}} 
& 
\textbf{Output (Expected)} \newline
\small \texttt{\{ "Period": "Q1 2024", "Revenue": \{value: 4.3, unit: "B"\}, "Revenue\_YoY\_Growth": "12\%", "Consensus\_Delta": \{value: 150, unit: "M"\} \}} \\

\textbf{Notes} \newline
Lacks structure; prone to inconsistent formatting and unit mismatches.
& 
\textbf{Notes} \newline
Improves consistency, reduces hallucinations, enables downstream parsing and SQL conversion. \\
\bottomrule
\end{tabularx}
\end{table}

\section{Ablation Study: Rule-based Validation}

As shown in Table~\ref{tab:ablation}, disabling rule-based validation leads to significantly higher unit and period errors.

\begin{table}[h]
\centering
\caption{Ablation: Without Rule-Based Validation}
\label{tab:ablation}
\begin{tabular}{lccc}
\toprule
\textbf{Metric} & \textbf{With Rules} & \textbf{Without Rules} & \textbf{$\delta$ Error} \\
\midrule
\makecell[l]{Extraction \\ Precision}     & 91.2\% & 82.6\% & -8.6\% \\
\midrule
\makecell[l]{Unit Error \\ Rate}          & 2.3\%  & 12.9\% & +10.6\% \\
\midrule
\makecell[l]{Period \\ Misalignment Rate} & 3.1\%  & 15.4\% & +12.3\% \\
\midrule
\makecell[l]{QA Match \\ Accuracy}        & 90.4\% & 78.7\% & -11.7\% \\
\midrule
\makecell[l]{SQL Validity \\ Pass Rate}   & 94.8\% & 86.0\% & -8.8\% \\
\bottomrule
\end{tabular}
\end{table}

\subsection*{Prompt A: Naive Zero-Shot Prompt (No Agent Iteration)}

\textbf{Prompt:}
\begin{quote}
"Extract all financial metrics mentioned in the following text and return them as a JSON object."
\end{quote}

\textbf{Example Input Text:}
\begin{quote}
"Operating margin in Q4 2024 was 14.6\%, up from 14.4\% last year. Revenue grew 15.2\% to \$2.52 billion. The company expects operating margin to be between 15–17\% in FY 2025."
\end{quote}

\textbf{LLM Output:}
```json
{
  "Operating margin": "14.6",
  "Revenue": "\$2.52 billion"
}

\bibliographystyle{ACM-Reference-Format}
\bibliography{main}

%%% -*-BibTeX-*-
%%% Do NOT edit. File created by BibTeX with style
%%% ACM-Reference-Format-Journals [18-Jan-2012].

\begin{thebibliography}{19}

%%% ====================================================================
%%% NOTE TO THE USER: you can override these defaults by providing
%%% customized versions of any of these macros before the \bibliography
%%% command.  Each of them MUST provide its own final punctuation,
%%% except for \shownote{} and \showURL{}.  The latter two
%%% do not use final punctuation, in order to avoid confusing it with
%%% the Web address.
%%%
%%% To suppress output of a particular field, define its macro to expand
%%% to an empty string, or better, \unskip, like this:
%%%
%%% \newcommand{\showURL}[1]{\unskip}   % LaTeX syntax
%%%
%%% \def \showURL #1{\unskip}           % plain TeX syntax
%%%
%%% ====================================================================

\ifx \showCODEN    \undefined \def \showCODEN     #1{\unskip}     \fi
\ifx \showISBNx    \undefined \def \showISBNx     #1{\unskip}     \fi
\ifx \showISBNxiii \undefined \def \showISBNxiii  #1{\unskip}     \fi
\ifx \showISSN     \undefined \def \showISSN      #1{\unskip}     \fi
\ifx \showLCCN     \undefined \def \showLCCN      #1{\unskip}     \fi
\ifx \shownote     \undefined \def \shownote      #1{#1}          \fi
\ifx \showarticletitle \undefined \def \showarticletitle #1{#1}   \fi
\ifx \showURL      \undefined \def \showURL       {\relax}        \fi
% The following commands are used for tagged output and should be
% invisible to TeX
\providecommand\bibfield[2]{#2}
\providecommand\bibinfo[2]{#2}
\providecommand\natexlab[1]{#1}
\providecommand\showeprint[2][]{arXiv:#2}

\bibitem[Agashe et~al\mbox{.}(2023)]%
        {agashe2023coordination}
\bibfield{author}{\bibinfo{person}{Shreyas Agashe}, \bibinfo{person}{Yifan Fan}, \bibinfo{person}{Aidan Reyna}, {and} \bibinfo{person}{Xiaoyuan~Edward Wang}.} \bibinfo{year}{2023}\natexlab{}.
\newblock \showarticletitle{LLM-Coordination: Evaluating and Analyzing Multi-Agent Coordination Abilities in Large Language Models}.
\newblock \bibinfo{journal}{\emph{arXiv preprint arXiv:2310.03903}} (\bibinfo{year}{2023}).
\newblock
\urldef\tempurl%
\url{https://arxiv.org/abs/2310.03903}
\showURL{%
\tempurl}


\bibitem[Costantino and Coletti(2008)]%
        {costantino2008information}
\bibfield{author}{\bibinfo{person}{Mario Costantino} {and} \bibinfo{person}{Paolo Coletti}.} \bibinfo{year}{2008}\natexlab{}.
\newblock \bibinfo{booktitle}{\emph{Information Extraction in Finance}}.
\newblock \bibinfo{publisher}{WIT Press}.
\newblock


\bibitem[Costantino and Morgan(1997)]%
        {costantino1997nlp}
\bibfield{author}{\bibinfo{person}{Mario Costantino} {and} \bibinfo{person}{R.~G. Morgan}.} \bibinfo{year}{1997}\natexlab{}.
\newblock \showarticletitle{Natural language processing and information extraction: Qualitative analysis of financial news articles}. In \bibinfo{booktitle}{\emph{Proceedings of the IEEE International Conference on Systems, Man, and Cybernetics}}. IEEE, \bibinfo{pages}{3870--3875}.
\newblock


\bibitem[Jacobs and Hoste(2022)]%
        {jacobs2022sentivent}
\bibfield{author}{\bibinfo{person}{Gilles Jacobs} {and} \bibinfo{person}{Veronique Hoste}.} \bibinfo{year}{2022}\natexlab{}.
\newblock \showarticletitle{SENTiVENT: enabling supervised information extraction of company-specific events in economic and financial news}.
\newblock \bibinfo{journal}{\emph{Language Resources and Evaluation}} \bibinfo{volume}{56}, \bibinfo{number}{4} (\bibinfo{year}{2022}), \bibinfo{pages}{1121--1150}.
\newblock


\bibitem[Kumar et~al\mbox{.}(2024)]%
        {kumar2024booksql}
\bibfield{author}{\bibinfo{person}{Raghav Kumar}, \bibinfo{person}{Akshay Dibbu}, \bibinfo{person}{Shubham Harsola}, {et~al\mbox{.}}} \bibinfo{year}{2024}\natexlab{}.
\newblock \showarticletitle{BookSQL: A Large Scale Text-to-SQL Dataset for Accounting Domain}.
\newblock \bibinfo{journal}{\emph{arXiv preprint arXiv:2401.12345}} (\bibinfo{year}{2024}).
\newblock
\urldef\tempurl%
\url{https://arxiv.org/abs/2401.12345}
\showURL{%
\tempurl}


\bibitem[Li et~al\mbox{.}(2024a)]%
        {li2024extracting}
\bibfield{author}{\bibinfo{person}{Hong Li}, \bibinfo{person}{Hao Gao}, {and} \bibinfo{person}{Chao Wu}.} \bibinfo{year}{2024}\natexlab{a}.
\newblock \showarticletitle{Extracting financial data from unstructured sources: Leveraging large language models}.
\newblock \bibinfo{journal}{\emph{Journal of Information Systems}} (\bibinfo{year}{2024}).
\newblock
\urldef\tempurl%
\url{https://publications.aaahq.org/jis/article/38/1/129/12345}
\showURL{%
\tempurl}


\bibitem[Li et~al\mbox{.}(2024b)]%
        {li2024survey}
\bibfield{author}{\bibinfo{person}{Xiaoxi Li}, \bibinfo{person}{Sheng Wang}, \bibinfo{person}{Shu Zeng}, \bibinfo{person}{Ying Wu}, {and} \bibinfo{person}{Yihan Yang}.} \bibinfo{year}{2024}\natexlab{b}.
\newblock \showarticletitle{A Survey on LLM-Based Multi-Agent Systems: Workflow, Infrastructure, and Challenges}.
\newblock \bibinfo{journal}{\emph{Springer}} (\bibinfo{year}{2024}).
\newblock
\urldef\tempurl%
\url{https://link.springer.com/article/10.1007/s10462-024-10621-2}
\showURL{%
\tempurl}


\bibitem[Malik et~al\mbox{.}(2011)]%
        {malik2011accurate}
\bibfield{author}{\bibinfo{person}{Hammad~H Malik}, \bibinfo{person}{Vishwas~S Bhardwaj}, {and} \bibinfo{person}{Herb Fiorletta}.} \bibinfo{year}{2011}\natexlab{}.
\newblock \showarticletitle{Accurate information extraction for quantitative financial events}. In \bibinfo{booktitle}{\emph{Proceedings of the 34th International ACM SIGIR Conference on Research and Development in Information Retrieval}}. \bibinfo{publisher}{ACM}, \bibinfo{pages}{1253--1254}.
\newblock


\bibitem[Shah et~al\mbox{.}(2024)]%
        {shah2024multi}
\bibfield{author}{\bibinfo{person}{Sarthak Shah}, \bibinfo{person}{Sai Ryali}, {and} \bibinfo{person}{Raghav Venkatesh}.} \bibinfo{year}{2024}\natexlab{}.
\newblock \showarticletitle{Multi-Document Financial Question Answering using LLMs}.
\newblock \bibinfo{journal}{\emph{arXiv preprint arXiv:2411.07264}} (\bibinfo{year}{2024}).
\newblock
\urldef\tempurl%
\url{https://arxiv.org/abs/2411.07264}
\showURL{%
\tempurl}


\bibitem[Sheikh and Conlon(2012)]%
        {sheikh2012rule}
\bibfield{author}{\bibinfo{person}{Mudassar Sheikh} {and} \bibinfo{person}{Sumali Conlon}.} \bibinfo{year}{2012}\natexlab{}.
\newblock \showarticletitle{A rule-based system to extract financial information}.
\newblock \bibinfo{journal}{\emph{Journal of Computer Information Systems}} \bibinfo{volume}{52}, \bibinfo{number}{4} (\bibinfo{year}{2012}), \bibinfo{pages}{88--97}.
\newblock


\bibitem[Singh et~al\mbox{.}(2025)]%
        {singh2025survey}
\bibfield{author}{\bibinfo{person}{Akshay Singh}, \bibinfo{person}{Aishwarya Shetty}, \bibinfo{person}{Atif Ehtesham}, {et~al\mbox{.}}} \bibinfo{year}{2025}\natexlab{}.
\newblock \showarticletitle{A Survey of Large Language Model-Based Generative AI for Text-to-SQL: Benchmarks, Applications, Use Cases, and Challenges}. In \bibinfo{booktitle}{\emph{Proceedings of the 2025 IEEE 15th International Conference on Big Data Science and Engineering (BDSE)}}. \bibinfo{publisher}{IEEE}.
\newblock


\bibitem[Song et~al\mbox{.}(2024)]%
        {song2024enhancing}
\bibfield{author}{\bibinfo{person}{Yujie Song}, \bibinfo{person}{Saif Ezzini}, \bibinfo{person}{Xia Tang}, \bibinfo{person}{Carl Lothritz}, {and} \bibinfo{person}{Janos Klein}.} \bibinfo{year}{2024}\natexlab{}.
\newblock \showarticletitle{Enhancing Text-to-SQL Translation for Financial System Design}. In \bibinfo{booktitle}{\emph{Proceedings of the 46th ACM Symposium on Applied Computing}}. \bibinfo{publisher}{ACM}.
\newblock


\bibitem[Srivastava et~al\mbox{.}(2024)]%
        {srivastava2024evaluating}
\bibfield{author}{\bibinfo{person}{Prashant Srivastava}, \bibinfo{person}{Meghna Malik}, \bibinfo{person}{Varun Gupta}, \bibinfo{person}{Tapan Ganu}, {et~al\mbox{.}}} \bibinfo{year}{2024}\natexlab{}.
\newblock \showarticletitle{Evaluating LLMs' Mathematical Reasoning in Financial Document Question Answering}.
\newblock \bibinfo{journal}{\emph{arXiv preprint arXiv:2402.13555}} (\bibinfo{year}{2024}).
\newblock
\urldef\tempurl%
\url{https://arxiv.org/abs/2402.13555}
\showURL{%
\tempurl}


\bibitem[Tran et~al\mbox{.}(2025)]%
        {tran2025multiagent}
\bibfield{author}{\bibinfo{person}{Khanh-Tung Tran}, \bibinfo{person}{Duy Dao}, \bibinfo{person}{Minh-Duy Nguyen}, \bibinfo{person}{Quoc-Viet Pham}, {et~al\mbox{.}}} \bibinfo{year}{2025}\natexlab{}.
\newblock \showarticletitle{Multi-Agent Collaboration Mechanisms: A Survey of LLMs}.
\newblock \bibinfo{journal}{\emph{arXiv preprint arXiv:2501.06322}} (\bibinfo{year}{2025}).
\newblock
\urldef\tempurl%
\url{https://arxiv.org/abs/2501.06322}
\showURL{%
\tempurl}


\bibitem[Wretblad and Riseby(2024)]%
        {wretblad2024bridging}
\bibfield{author}{\bibinfo{person}{Niklas Wretblad} {and} \bibinfo{person}{Fredrik~Gordh Riseby}.} \bibinfo{year}{2024}\natexlab{}.
\newblock \showarticletitle{Bridging Language and Data: Optimizing Text-to-SQL Generation in Large Language Models}.
\newblock \bibinfo{journal}{\emph{DIVA Portal}} (\bibinfo{year}{2024}).
\newblock
\urldef\tempurl%
\url{https://www.diva-portal.org/smash/get/diva2:1823187/FULLTEXT01.pdf}
\showURL{%
\tempurl}


\bibitem[Yang et~al\mbox{.}(2024)]%
        {yang2024evaluating}
\bibfield{author}{\bibinfo{person}{Xinqi Yang}, \bibinfo{person}{Scott Zang}, \bibinfo{person}{Yong Ren}, \bibinfo{person}{Dingjie Peng}, {and} \bibinfo{person}{Zheng Wen}.} \bibinfo{year}{2024}\natexlab{}.
\newblock \showarticletitle{Evaluating Large Language Models on Financial Report Summarization: An Empirical Study}.
\newblock \bibinfo{journal}{\emph{arXiv preprint arXiv:2411.06852}} (\bibinfo{year}{2024}).
\newblock
\urldef\tempurl%
\url{https://arxiv.org/abs/2411.06852}
\showURL{%
\tempurl}


\bibitem[Yang et~al\mbox{.}(2025)]%
        {yang2025agentnet}
\bibfield{author}{\bibinfo{person}{Yujing Yang}, \bibinfo{person}{Haotian Chai}, \bibinfo{person}{Shu Shao}, \bibinfo{person}{Yihan Song}, \bibinfo{person}{Shan Qi}, \bibinfo{person}{Ruijie Rui}, {et~al\mbox{.}}} \bibinfo{year}{2025}\natexlab{}.
\newblock \showarticletitle{AgentNet: Decentralized Evolutionary Coordination for LLM-Based Multi-Agent Systems}.
\newblock \bibinfo{journal}{\emph{arXiv preprint arXiv:2503.10777}} (\bibinfo{year}{2025}).
\newblock
\urldef\tempurl%
\url{https://arxiv.org/abs/2503.10777}
\showURL{%
\tempurl}


\bibitem[Zhang et~al\mbox{.}(2024)]%
        {zhang2024finsql}
\bibfield{author}{\bibinfo{person}{Cheng Zhang}, \bibinfo{person}{Yifan Mao}, \bibinfo{person}{Yu Fan}, \bibinfo{person}{Yutong Mi}, \bibinfo{person}{Yu Gao}, {and} \bibinfo{person}{Ling Chen}.} \bibinfo{year}{2024}\natexlab{}.
\newblock \showarticletitle{FinSQL: Model-Agnostic LLMs-Based Text-to-SQL Framework for Financial Analysis}.
\newblock \bibinfo{journal}{\emph{Companion of the 2024 ACM International Conference on AI in Finance}} (\bibinfo{year}{2024}).
\newblock


\bibitem[Ziegler(2024)]%
        {ziegler2024automating}
\bibfield{author}{\bibinfo{person}{Guilherme~Gomes Ziegler}.} \bibinfo{year}{2024}\natexlab{}.
\newblock \showarticletitle{Automating Information Extraction from Financial Reports Using LLMs}.
\newblock \bibinfo{journal}{\emph{Aalto University Publications}} (\bibinfo{year}{2024}).
\newblock
\urldef\tempurl%
\url{https://aaltodoc.aalto.fi/handle/123456789/123456}
\showURL{%
\tempurl}


\end{thebibliography}

\end{document}